\documentclass[conference]{IEEEtran}
\ifCLASSINFOpdf
\else
\fi
\usepackage[cmex10]{amsmath}
\makeatletter
\let\IEEEcaption\@makecaption
\makeatother

\usepackage[font=footnotesize]{subcaption}

\makeatletter
\let\@makecaption\IEEEcaption
\makeatother

\usepackage{graphicx}

\usepackage{tikz}
\usetikzlibrary{calc,intersections,decorations}
\usepackage{pgfplots}
\pgfplotsset{compat=1.8}

\pgfmathdeclarefunction{gauss}{2}{%
	\pgfmathparse{exp(-((x-#1)^2)/(2*#2^2))}%
}
\pgfplotsset{
 	axis line style={black}
}

\usepackage[%
style=numeric,%
citestyle=numeric,%
bibstyle=numeric,%
useprefix=false,%
maxbibnames=99,%
uniquename=false,%
firstinits=true,%
isbn=false,%
doi=false,%
sorting=ayt,%
backend=biber,%
sortcites=true]{biblatex}

\DeclareSortingScheme{ayt}
{
	\sort{
		\field{presort}
	}
	\sort[final]{
		\field{sortkey}
	}
	\sort{
		\name{sortname}
		\name{author}
		\name{editor}
		\name{translator}
		\field{sorttitle}
		\field{title}
	}
	\sort{
		\field{sortyear}
		\field{year}
	}
	\sort{
		\field{sorttitle}
		\field{title}
	}
}

\DeclareSortingScheme{yat}
{
	\sort{
		\field{presort}
	}
	\sort[final]{
		\field{sortkey}
	}
	\sort{
		\field{sortyear}
		\field{year}
	}
	\sort{
		\name{sortname}
		\name{author}
		\name{editor}
		\name{translator}
		\field{sorttitle}
		\field{title}
	}
	\sort{
		\field{sorttitle}
		\field{title}
	}
}

\DeclareNameAlias{sortname}{last-first}
\DeclareNameAlias{default}{last-first}

\DefineBibliographyStrings{english}{%
	bibliography = {References},
}

\usepackage{xpatch}

\xpatchbibmacro{name:andothers}{ 
	{\finalandcomma}%
}{%
\addspace%
}{}{}

\usepackage[parfill]{parskip}

\renewbibmacro{in:}{}

\usepackage{chngcntr}
\counterwithout{footnote}{section}

\usepackage[bottom]{footmisc}

\title{Adaptive conversion of real-valued input\\into spike trains}

\author{\IEEEauthorblockN{Alexander Hadjiivanov}
\IEEEauthorblockA{School of Computer Science and Engineering\\
University of New South Wales\\
Sydney, NSW 2052, Australia \\
Email: a.hadjiivanov@student.unsw.edu.au}}

\begin{document}

\maketitle

\begin{abstract}
This paper presents a biologically plausible method for converting real-valued input into spike trains for processing with spiking neural networks.
The proposed method mimics the adaptive behaviour of retinal ganglion cells and allows input neurons to adapt their response to changes in the statistics of the input.
Thus, rather than passively receiving values and forwarding them to the hidden and output layers, the input layer acts as a self-regulating filter which
emphasises deviations from the average while allowing the input neurons to become effectively desensitised to the average itself.
Another merit of the proposed method is that it requires only one input neuron per variable, rather than an entire population of neurons as
in the case of the commonly used conversion method based on Gaussian receptive fields.
In addition, since the statistics of the input emerge naturally over time, it becomes unnecessary to pre-process the data before feeding it to the network.
This enables spiking neural networks to process raw, non-normalised streaming data.
A proof-of-concept experiment is performed to demonstrate that the proposed method operates as expected.
\end{abstract}


%
\IEEEpeerreviewmaketitle

\section{Introduction}
Spiking neural networks (NNs) have been shown to have considerably higher processing power than classical NNs with a sigmoidal activation function \parencite{Maass--1996}.
Although certain types of classical NNs (such as recurrent NNs) can process temporal sequences, in general they are incapable of
making use of temporal correlations in the input.
In contrast, spiking NNs can harness their inherent sensitivity to time to discover such temporal correlations by taking into account the relative timings between spikes.
However, while various mechanisms, applications and implementations for spike-based processing have been studied in depth \parencite{Maass--1995,Maass--1997,Izhikevich--2000,Izhikevich--2004,Paugam-Moisy--2006,Kasabov_Dhoble_Nuntalid_Indiveri--2013,Bohte_Kok_LaPoutre--2002,Delorme_Thorpe--2001},
surprisingly little attention has been paid to the problem of converting
real-valued input into spike trains. Note that this is separate to the issue of \textit{encoding} information using spike trains, which has been
	investigated in detail (e.g., see \parencite{Thorpe_Delorme_Rullen--2001} for a survey of encoding techniques).
The most commonly used input conversion method utilises a population of input neurons with overlapping Gaussian sensitivity profiles
(Gaussian receptive fields; GRF below) to convert real-valued input into spike latencies.
An overview of the GRF method is given in Section \ref{previous_work}.

There are several issues associated with GRF-based input conversion. First, each variable is mapped to a population of neurons rather than a single neuron.
There is no rigorous technique for determining the optimal number and density of neurons, and therefore they are chosen arbitrarily depending on the application.
This results in considerable redundancy because often only a fraction of the input neurons will fire when a data point is presented, and the remainder of the population will remain silent.

Another issue concerns the range of the input variables. GRF-based conversion requires that the minimum and maximum of each variable be known before the network is trained
(indeed, before it is even designed) in order to decide how many input neurons to allocate to each population.
However, extremal values may not be available in advance, such as in the case of processing certain types of streaming data.
The GRF method could still be used in such cases if each population of neurons is mapped \textit{a priori} to the interval $[0,1]$.
In this case, all incoming data points must be normalised before being presented to the network.
However, this can be computationally taxing or outright infeasible if the data inflow rate is too high.

Finally, the response of a neuron with a GRF is deterministic, meaning that it always outputs the same latency for a given input value.
This prevents the network from adapting to changes in the statistics of the input, which provides another obstacle to the application of spiking NNs to streaming data.

In short, GRF-based conversion is rigid and prone to redundancy.
This paper introduces a novel, biologically plausible adaptive method for converting real-valued input into spike trains.
The proposed method addresses all of the above problems and offers a computational advantage over the GRF method combined with high flexibility and natural applicability to streaming data.


\section{Previous Work}\label{previous_work}
GRF-based conversion is the most commonly used method for converting real-valued input into spike trains.
The principle of this method is briefly outlined below.

Each input variable $x_{m}$ is captured by a population of input neurons. It is necessary to know in advance the corresponding minimal and maximal values ($x^{min}_{m}, x^{max}_{m}$),
in other words, the range of possible values for that variable.
Note that this range is fixed and cannot be changed once the network becomes operational.
The range of variable $x_{m}$ can be covered by $N$ input neurons with Gaussian sensitivity profiles arranged as shown in Fig. \ref{fig:grf}.
The average value and the standard deviation of the sensitivity profile of input neuron $i = 1, ..., N$ are calculated as follows:

\begin{figure}
\centering

	\begin{tikzpicture}
		\tikzset{neuron/.style={
				shape=circle,
				minimum size=1mm,
				inner sep=0,
				fill=black}
		}
		\newcommand{\gaussplot}[3]{%
			\addplot
			[%
			name path global=gaussian#1,
			mark=none,
			samples=100,
			smooth,
			thin,
			color=gray
			]
			{1 - gauss(#1,#2)};
			\node[neuron,below,label={[font=\tiny]below:#3}] (neuron#1) at (axis cs: #1,0){};
			\path [name path global=v#1] (neuron#1) -- (axis cs: #1,1);
		}
		\begin{axis}%
		[%
		axis x line=center,
		axis y line=left,
		ytick={0,0.85},
		xtick={},
		xticklabels={},
		yticklabels={0,$\theta$},
		ymin=0,
		clip=false,
		ymax=1.4,
		yscale=0.55,
		xlabel=$x_{m}$,
		xlabel style={anchor=north},
		ylabel=Spike latency,
		domain=-4:4
		]
		\gaussplot{-3}{0.75}{$n_{1}$};
		\gaussplot{-2}{0.75}{};
		\gaussplot{-1}{0.75}{};
		\gaussplot{0}{0.75}{$n_{i}$};
		\gaussplot{1}{0.75}{$n_{i+1}$};
		\gaussplot{2}{0.75}{$n_{i+2}$};
		\gaussplot{3}{0.75}{$n_{i+3}$};
		\draw[name path=xmt,dashed,teal] (axis cs: 1.40,1) -- (axis cs:1.40,0);
		\draw[dashed] (axis cs: -3,0) -- (axis cs:-3,1);
		\draw[dashed] (axis cs: 3,0) -- (axis cs:3,1);
		\node[text=black,above] at (axis cs: -3,1) {$x_{m}^{min}$};
		\node[text=black,above] at (axis cs: 3,1) {$x_{m}^{max}$};
		\node[text=black,above] at (axis cs: 1.4,1) {$x_{m}(t)$};
		\addplot[mark=none,blue,name path=threshold,densely dotted] {0.85};

		\path [name intersections={of=xmt and gaussian0,by=l0}];
		\path [name intersections={of=xmt and gaussian1,by=l1}];
		\path [name intersections={of=xmt and gaussian2,by=l2}];
		\path [name intersections={of=xmt and gaussian3,by=l3}];

		\node[fill=red,inner sep=1pt] at (l0) {};
		\node[fill=red,inner sep=1pt] at (l1) {};
		\node[fill=red,inner sep=1pt] at (l2) {};
		\node[fill=red,inner sep=1pt] at (l3) {};

		\draw[red,densely dotted,name path=edge0] (l0) -| (neuron0);
		\draw[red,densely dotted,name path=edge1] (l1) -| (neuron1);
		\draw[red,densely dotted,name path=edge2] (l2) -| (neuron2);
		\draw[red,densely dotted,name path=edge3] (l3) -| (neuron3);

		\path [red,name intersections={of=v0 and edge0,by=corner0}];
		\path [red,name intersections={of=v1 and edge1,by=corner1}];
		\path [red,name intersections={of=v2 and edge2,by=corner2}];
		\path [red,name intersections={of=v3 and edge3,by=corner3}];

		\draw[red,decorate,decoration={brace,mirror}] (corner0) -- node[anchor=east,midway,text=black] {\tiny $\lambda_{i}$} (neuron0);
		\draw[red,decorate,decoration={brace,mirror}] (corner1) -- node[anchor=east,midway,text=black] {\tiny $\lambda_{i+1}$} (neuron1);
		\draw[red,decorate,decoration=brace] (corner2) -- node[anchor=west,midway,text=black] {\tiny $\lambda_{i+2}$} (neuron2);
		\draw[red,decorate,decoration=brace] (corner3) -- node[anchor=west,midway,text=black] {\tiny $\lambda_{i+3}$} (neuron3);

		\end{axis}
	\end{tikzpicture}

	\caption[GRF-based conversion of real-valued input into spike latencies]{GRF-based method for converting a real-valued input variable into spike latencies.
		Input neurons are placed at the peaks of inverted Gaussians, the input variable $x_{m}$ is mapped to the $x$ axis, and the spike latency increases along the $y$ axis.
		The latency $\lambda_{i}$ for neuron $n_{i}$ is computed as the vertical offset from the peak of the corresponding Gaussian and the point
		where the vertical line passing through the value of $x_{m}$ at timestep $t$ (denoted as $x_{m}(t)$) crosses that Gaussian.
		If the latency is greater than a certain threshold $\theta$, it is considered that the neuron does not produce a spike.
		In the example above, $\lambda_{i+1} < \lambda_{i+2} < \lambda_{i} $, and therefore neuron $n_{i+1}$ spikes first, followed by neurons $n_{i+2}$
		and $n_{i}$. Neuron $n_{i+3}$ does not produce a spike because the corresponding latency $\lambda_{i+3}$ is greater than the threshold $\theta$.}
	\label{fig:grf}
\end{figure}
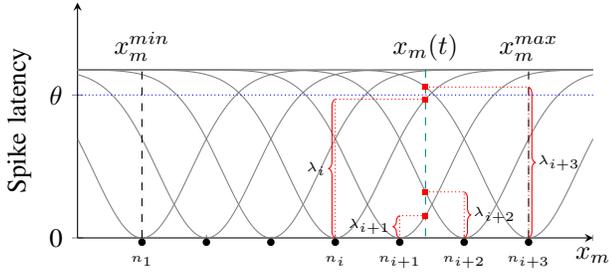

\begin{equation}\label{eq:GRF_mu}
\mu_{i} = x^{min}_{m} + \frac{(2i - 3)(x^{max}_{m} - x^{min}_{m})}{2(N - 2)}
\end{equation}

\begin{equation}\label{eq:GRF_sigma}
\sigma = \frac{x^{max}_{m} - x^{min}_{m}}{\beta(N - 2)}
\end{equation}

Here, $\mu_{i}$ is the mean of the $i$-th Gaussian profile, $\sigma$ is the standard deviation of the Gaussian profiles (the same for all neurons),
$\beta$ is a parameter which controls the width of the Gaussian profiles, and $N$ is the number of neurons in the population.

An early prototype of this method was studied in detail in the context of distributed representations as a possible approach to modelling associative memory \parencite{Hinton_McClelland_Rumelhart--1986}.
Arguably, one advantage of using overlapping receptive fields is that similar values would tend to evoke similar responses from the population of neurons,
which is indeed a desirable property for memory applications.

GRF-based conversion of real-valued input in its current form was introduced as an idealisation of a `neural map' capable of capturing
the input with high resolution by using a population of low-resolution receptive fields \parencite{Heiligenberg--1987}.
It has been demonstrated that for sufficiently broad tuning of the sensitivity profiles, the response becomes a linear function
of the input. However, the linearity breaks at the ends of the neuron array \parencite{Heiligenberg--1987,Baldi_Heiligenberg--1988}.

The biological structures which have inspired the GRF-based conversion method are suitable for representing the \textit{spatial} properties of the input.
In particular, overlapping GRFs provide a plausible explanation of the phenomenon of hyperacuity \parencite{Westheimer--2009}, which is
ubiquitous in nature and can be seen in the visual \parencite{Shapley_Victor--1986}, auditory \parencite{Simmons--1989} and possibly other sensory systems in various animals.
An array of GRFs ensures that the \textit{instantaneous} magnitude of the input is captured with high resolution.
However, it cannot give any information about how it changes with time.

In this regard, it has been shown that the response of certain receptor neurons in biological sensory systems
adapts dynamically to changes in the surrounding environment.
The most striking example of such adaptation is provided by retinal ganglion cells, whose response characteristics change depending on the recent history of illumination of the retina
in a manner resembling automatic gain control \parencite{Smirnakis_Berry_Warland_Bialek_Meister--1997,Kaplan_Benardete--2001,Cleland_Enroth-Cugell--1970,Laughlin_Hardie--1978}.
It has been demonstrated that at different luminance levels, the response curve of retinal ganglion cells at different average luminance levels has virtually the same shape but a shifted
mean, which allows the cells to maintain an optimal dynamic range at any given average luminance level \parencite{Barlow_Foldiak--1989}.
This greatly reduces the amount of information which has to be conveyed to the visual cortex at any given time because
the receptors transmit only the intensity of the stimulus relative to the background rather than the absolute intensity.
Thus, retinal ganglion cells achieve highly efficient encoding of information by transmitting only those parts of the input
which are necessary to detect \textit{deviations} from the average intensity of the input \parencite{Barlow--1961}.
Similar adaptation also occurs after exposing the eye to various patterns while the eye is fixated on a single point, giving rise to after-images.
This mechanism has been suggested as a possible explanation of optical illusions \parencite{Barlow--1990}.
It has even been suggested that such adaptation is a form of \textit{predictive} coding aimed at filtering out the most predictable information in the
environment in order to pass the remaining (less predictable) information to the central nervous system for further processing \parencite{Hosoya_Baccus_Meister--2005}.
In analysing the adaptability of the retina, \textcite[p. 72]{Smirnakis_Berry_Warland_Bialek_Meister--1997} conclude that ``[...] it might be expected that any neural circuit would
benefit from an adaptive control that responds to changes in the statistics of its inputs''.

In light of the above, the following section introduces a model of adaptive input conversion for spiking NNs
which aims to mimic the adaptive response of retinal ganglion cells to changes in the intensity of environmental stimuli.

\section{Model of Adaptive Input Conversion}

\subsection{Keeping Track of the Average}
As mentioned the preceding section, the response of the retina (and possibly other sensory systems) adapts to the average intensity of the stimulus.
The goal of this study is to mimic this behaviour in order to convert arbitrary real-valued input variables whose underlying statistics are unknown into coherent spike trains.

The first step is to employ a mechanism capable of accumulating the statistics on the fly.
This can be easily achieved by keeping track of the \textbf{running average} and \textbf{running variance} of each variable.
Assuming that variable $x$ changes with time, its value can be sampled with some (arbitrary) sampling rate.
The value $x_{i}$ arriving at step $i$ can be used to update the running average $\mu_{i}$ and the running (sample) variance $s_{i}$ as follows:

\begin{equation}\label{eq:running_average}
\mu_{i} =	\begin{cases}
					x_{i}, & i = 1\\
					\mu_{i - 1} + \frac{x_{i} - \mu_{i}}{i}, & i > 1
				\end{cases},
\end{equation}

\begin{equation}\label{eq:running_variance}
s_{i} =	\begin{cases}
				0, & i = 1\\
				s_{i - 1} + (x_{i} - \mu_{i - 1})(x_{i} - \mu_{i}), & i > 1
			\end{cases}
\end{equation}

Naturally, the first value $x_{1}$ does not provide any meaningful statistics about the variable. Therefore, as a boundary condition at the first step, the average is set to the
first value and the variance is set to $0$. As new values arrive, the average and the variance gradually change and converge to the true average and standard deviation of the variable distribution.
If the statistics of the input change in any way, the running average will track this change by
moving in a direction that effectively minimises the difference between the current value and the average.

\subsection{Conversion Into Spike Trains}
The next step is to translate the input into a spike train.
In the GRF-based method, the input variable is mapped to the $x$ axis, and the GRFs convert the current value into spike latencies \parencite{Schliebs_Kasabov--2013}.
Neurons located closer to the current value spike earlier, whereas ones which are further away spike later or not at all.

\begin{figure}
	\centering

	\begin{tikzpicture}[trim axis left,trim axis right]
	\begin{axis}%
	[%
	axis x line=bottom,
	axis y line=left,
	ymin=0,
	xmin=0,
	ymax=1.2,
	xmax=15,
	yscale=0.5,
	ytick={1},
	yticklabels={$\theta$},
	xtick={4,6.5,12},
	xticklabels={$\lambda(I_{2})$,$\lambda(I_{1})$,$\lambda(I_{0})$},
	xlabel=Stimulus detection latency,
	xlabel style={below=2mm},
	ylabel=Depolarisation level
	]

	\addplot
	[%
	mark=none,
	thin,
	name path=threshold,
	dotted,
	domain=0:15
	] {1};

	\draw[black!60!green] (axis cs: 2,0) -- (axis cs: 4,1);
	\draw[dashed,red] (axis cs: 2,0) -- (axis cs: 6.5,1);
	\draw[dotted,blue] (axis cs: 2,0) -- (axis cs: 12,1);

	\draw[black,thick,dotted] (axis cs: 4,1) -- (axis cs: 4,0);
	\draw[black,thick,dotted] (axis cs: 6.5,1) -- (axis cs: 6.5,0);
	\draw[black,thick,dotted] (axis cs: 12,1) -- (axis cs: 12,0);

	\end{axis}

	\end{tikzpicture}

	\begin{tikzpicture}[trim axis left,trim axis right]
	\begin{axis}%
	[%
	axis x line=bottom,
	axis y line=left,
	ymin=0,
	ymax=1.2,
	yscale=0.5,
	ytick={0.3,0.6,0.9},
	yticklabels={$I_{0}$,$I_{1}$,$I_{2}$},
	xtick=\empty,
	xticklabels=\empty,
	xlabel=Time,
	ylabel=Stimulus intensity,
	domain=0:15
	]

	\addplot
	[%
	mark=none,
	domain=0:15
	] {0.0};

	\draw
	[%
	dotted,
	mark=none
	] (axis cs: 0,0.3) -- (axis cs: 2,0.3);

	\draw
	[%
	dotted,
	mark=none
	] (axis cs: 0,0.6) -- (axis cs: 2,0.6);

	\draw
	[%
	dotted,
	mark=none
	] (axis cs: 0,0.9) -- (axis cs: 2,0.9);

	\draw
	[%
	blue,
	thick,
	dotted,
	mark=none
	] (axis cs: 2,0) |- (axis cs: 15,0.3);

	\draw
	[%
	red,
	thick,
	dashed,
	mark=none
	] (axis cs: 2,0) |- (axis cs: 15,0.6);

	\draw
	[%
	black!50!green,
	thick,
	mark=none
	] (axis cs: 2,0) |- (axis cs: 15,0.9);

	\end{axis}
	\end{tikzpicture}

	\caption[A highly idealised representation of the relation between stimulus intensity and detection time]{A highly idealised representation of the relation between stimulus intensity and detection time.
		A strong stimulus results in faster depolarisation of the neuron membrane, thus shortening the time until the firing threshold is reached.
		If the stimulus is sustained, it will keep depolarising the neuron membrane, causing the neuron to fire repeatedly.
		However, due to the non-linear response characteristics of the neuron, doubling the stimulus intensity does not necessarily halve the response time (and thus the detection latency).}
	\label{fig:stimulus_detection_time}
\end{figure}
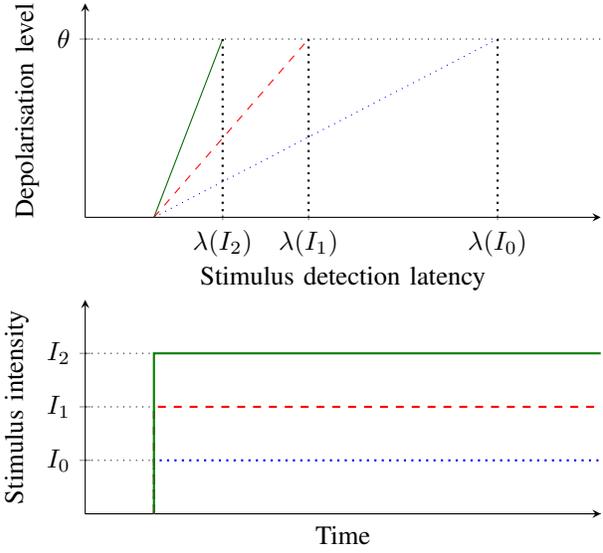

\begin{figure}
	\centering
	\subcaptionbox{\label{fig:RGC_response}}
	{
		\begin{tikzpicture}
		\begin{axis}%
		[%
		axis x line=center,
		axis y line=center,
		yscale=0.7,
		xtick={-1,01},,
		ytick={0,1},
		ymax=1.3,
		xmin=-3,
		xmax=3,
		xlabel=$x$,
		xlabel style={at={(current axis.right of origin)},anchor=north},
		legend style={at={(1.0,0.5)},anchor=east},
		yticklabel style={anchor=south east}
		]
		\addplot%
		[%
		red,
		dashed,
		mark=none,
		samples=1000,
		smooth,
		domain=-3:3,
		thick
		]
		(x,{1 - 1/(1+exp(-x))});
		\addplot%
		[%
		blue,
		mark=none,
		samples=1000,
		smooth,
		domain=-3:3
		]
		(x,{1/(1+exp(-x))});
		\legend{\tiny$1 - \sigma(x)$,\tiny$\sigma(x)$}

		\addplot
		[%
		black,
		mark=none,
		samples=2,
		domain=-3:3,
		very thin,
		dotted
		]
		{1};
		\end{axis}
		\end{tikzpicture}
	}
	\vspace{1.2em}
	\subcaptionbox{\label{fig:abs_tanh}}
	{
		\begin{tikzpicture}
		\begin{axis}%
		[
		axis x line=center,
		axis y line=center,
		yscale=0.7,
		xtick={-1,0,1},,
		ytick={-1,0,1},
		ymax=1.3,
		xmin=-3,
		xmax=3,
		xlabel=$x$,
		xlabel style={at={(current axis.right of origin)},anchor=south, below=7.5mm},
		legend style={at={(1.0,0.9)},anchor=east},
		yticklabel style={anchor=south east}
		]
		\addplot%
		[%
		red,
		dashed,
		mark=none,
		samples=1000,
		smooth,
		domain=-3:3,
		thick
		]
		(x,{abs(tanh(x))});
		\addplot%
		[%
		blue,
		mark=none,
		samples=1000,
		smooth,
		domain=-3:3
		]
		(x,{tanh(x)});
		\legend{\tiny$\left\lvert{\tanh(x)}\right\rvert$,\tiny$\tanh(x)$}

		\addplot
		[%
		black,
		mark=none,
		samples=2,
		domain=-3:3,
		very thin,
		dotted
		]
		{1};
		\addplot
		[
		black,
		mark=none,
		samples=2,
		domain=-3:3,
		very thin,
		dotted
		]
		{-1};
		\end{axis}
		\end{tikzpicture}
	}
	\vspace{1.2em}
	\subcaptionbox{\label{fig:receptive_field}}
	{
		\begin{tikzpicture}
		\begin{axis}%
		[
		axis x line=bottom,
		axis y line=center,
		yscale=0.7,
		xmin=-3,
		xmax=3,
		xtick={0},
		ytick={0,1},
		ymax=1.25,
		xlabel=$x$,
		xlabel style={at={(current axis.right of origin)},anchor=north},
		ylabel={$\rho(x_{i}) = \left\lvert{\tanh(\frac{x_{i} - \mu_{i}}{sd_{i}})}\right\rvert$},
		xticklabels={$\mu_{i}$},
		legend style={at={(1.0,0.3)},anchor=east},
		yticklabel style={anchor=south east}
		]
		\addplot%
		[%
		black!50!green,
		mark=none,
		samples=1000,
		smooth,
		domain=-3:3,
		]
		(x,{abs(tanh(x))});

		\addplot%
		[%
		red,
		dashed,
		mark=none,
		samples=1000,
		smooth,
		domain=-3:3,
		]
		(x,{abs(tanh(x/2))});

		\addplot%
		[%
		blue,
		dotted,
		mark=none,
		samples=1000,
		smooth,
		domain=-3:3,
		thick
		]
		(x,{abs(tanh(2*x))});

		\legend{\tiny$sd_{i} = 1$,\tiny$sd_{i} > 1$,\tiny$sd_{i} < 1$}

		\addplot
		[
		black,
		mark=none,
		samples=2,
		domain=-6:6,
		very thin,
		dotted
		]
		{1};
		\end{axis}
		\end{tikzpicture}
	}

	\caption[$\sigma(x)$, $\tanh(x)$ and $\rho(x)$]{(a) The logistic function $\sigma(x)$ (continuous) and the function $1-\sigma(x)$ (dashed), which is a mirror image of $\sigma(x)$ about the $y$ axis.
		These two functions can be used to approximate the respective response of on- and off-type cells in the retina.
		(b) The $\tanh$ function (continuous), which is commonly used as an activation function in classical NNs,
		plotted together with its absolute value (dashed).
		(c) The function in Eq. \ref{eq:spiking_rate_tanh} converting real-valued input into a spiking rate $\rho(x)$.
		Note that $\rho(x)$ is centred at $\mu_{i}$ rather than $0$,
		which provides a way to adapt to changes in the statistics of the input.
		The slope of the function at $\mu_{i}$ is indicative of the standard deviation of $x$. If $x$ is fairly stable (i.e., has a small standard deviation),
		the slope will be steep, and the transfer function will be sensitive to small deviations from the mean in either direction.
		In contrast, a wildly fluctuating input can be captured with $\rho(x)$ with a gentler slope to reflect the larger variance.
		This allows the network to perceive \textit{contrast} while
		remaining mostly insensitive to values close to the average value (`the background').
		Although the range of possible values of $\rho(x)$ is $(0,1)$, it can be multiplied by an arbitrary constant $\rho_{max}$ representing
		the maximal possible firing rate, thus expanding the firing rate range to the interval $(0,\rho_{max})$.}
	\label{fig:sigmoid_and_spike_rate}
\end{figure}
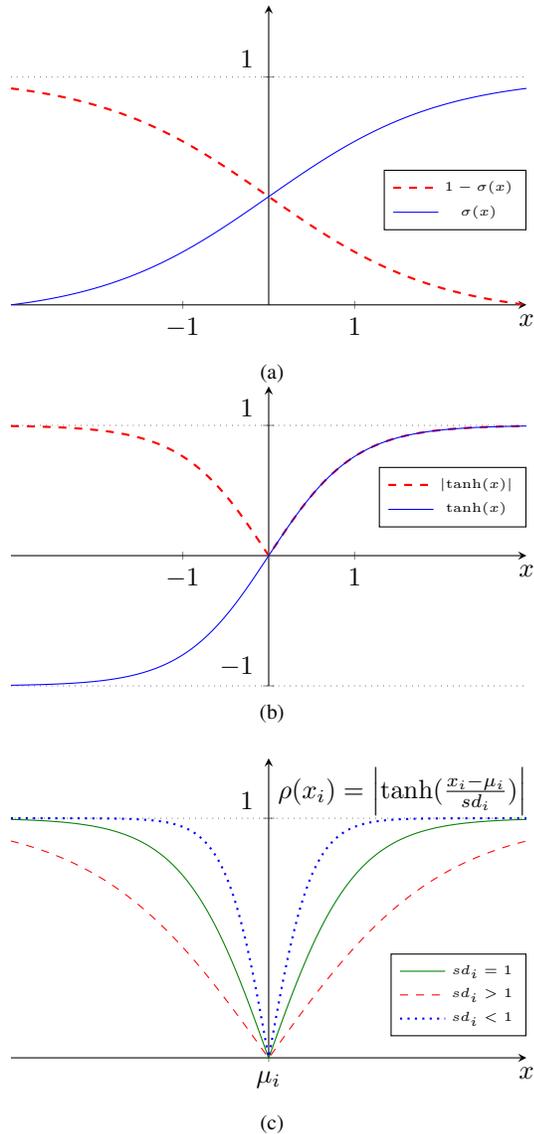

It is generally considered that sensory neurons respond to a stimulus with a latency inversely proportional to the intensity of the stimulus \parencite{Lennie--1981,Levick--1973}.
Idealised input conversion models, including the GRF-based model, produce one spike per neuron with an appropriate latency.
However, in practice the latency of stimulus detection is measured based on the amount of time required for the receptor to produce a certain number of spikes.
Generally, this interval is shorter for stronger stimuli because strong excitation of the receptor results in a higher spiking rate.
A possible explanation of this mechanism (without necessarily claiming biological plausibility) is shown in Fig. \ref{fig:stimulus_detection_time}.
It is therefore necessary to capture the directly proportional relation between stimulus intensity and firing rate to convert real-valued input into spike trains.
The hyperbolic tangent ($\tanh$) function (Fig. \ref{fig:abs_tanh}) is adopted for this purpose, similarly to classical NNs, where it is used precisely as an abstraction
representing the average firing rate of neurons.
However, the proposed model uses the absolute value of $\tanh$ because the spiking rate cannot be negative:

\begin{equation}\label{eq:spiking_rate_prop}
	\rho(x) \propto \left\lvert{\tanh(x)}\right\rvert,
\end{equation}

Note that $\rho(x)$ as defined above ensures that the spiking rate of the neuron increases non-linearly as the input value moves further away from $0$ in either direction,
as in the case of biological receptor neurons.

The rationale behind Eq. \ref{eq:spiking_rate_prop} is as follows.
The response characteristics of on/off-type cells in the retina are illustrated in Fig. \ref{fig:RGC_response} following the discussion presented in \parencite[Fig. 4.2]{Barlow_Foldiak--1989}.
If only the logistic function $\sigma(x) = \frac{1}{1+e^{-x}}$ (continuous plot in Fig. \ref{fig:RGC_response}) is used as a transfer function, input values which are lower than $0$ will result in a low spiking rate.
However, since we are interested in capturing departures from the average in either direction
(lower \textit{and} higher), it makes sense to model the neuron receptive field by using $\sigma(x)$ to capture values larger than $0$ and $1 - \sigma(x)$ to capture values smaller than $0$.
From a biological point of view, $\sigma(x)$ represents the response of on-type cells and $1 - \sigma(x)$ that of off-type cells.
In this way, the the spiking rate can be computed as the absolute difference of the two functions, as follows:

\begin{equation}\label{eq:RGC_response_model}
	\rho(x) = \left\lvert{\sigma(x) - (1-\sigma(x))}\rvert= \lvert{2\sigma(x) - 1}\right\rvert,
\end{equation}

In addition, whereas in classical NNs the sigmoid function is centred at $0$, in the proposed method the sigmoid is centred at the current average $\mu_{i}$
in order to take into account the accumulated knowledge about the statistics of the input.
Essentially, this defines a mechanism for adapting the dynamic range in a way that ensures optimal response regardless of the current average value of the input.

At step $i$, the input neuron is presented with a new value $x_{i}$, which is used to update the running average and to calculate the spiking rate.
It should be noted that the conversion must remain efficient over a reasonable range of values for $x$, which is the reason for keeping track of the running variance
as well as the running average.
Thus, the expression for the spiking rate at step $i$ takes the following form:

\begin{equation}\label{eq:spiking_rate_sigma}
	\rho(x_{i}) = \left\lvert{2\sigma({\frac{x_{i} - \mu_{i}}{sd_{i}}}) - 1}\right\rvert,
\end{equation}

where $sd_{i} = \sqrt{s_{i}}$ is the running standard deviation.
In this way, a large variance results in a `stretched' transfer function which is capable of accommodating a highly fluctuating variable.
If the standard deviation is small, the resulting transfer function will be steep in order to remain sensitive to small changes in $x$.

Finally, given the relation $2\sigma(x) - 1 = \tanh(\frac{x}{2})$, $\sigma(x)$ in Eq. \ref{eq:spiking_rate_sigma} can be replaced with a simple $\tanh$ function.
Assuming that the denominator $2$ is absorbed by the standard deviation $sd_{i}$, we arrive at the final equation for computing the spiking rate (Fig. \ref{fig:receptive_field}):

\begin{equation}\label{eq:spiking_rate_tanh}
	\rho(x_{i}) = \left\lvert{\tanh(\frac{x_{i} - \mu_{i}}{sd_{i}})}\right\rvert.
\end{equation}

\subsection{Controlling the Speed of Adaptation}
When accumulating the running statistics, the running average $\mu_{i}$ will become increasingly resilient to change because the contribution of new values will become progressively smaller
as $i$ grows. This obstacle can be eliminated by applying a window covering only the last $k$ data points.
For $i < k$, the running average and the running variance are computed as shown in Eqs. \eqref{eq:running_average} and \eqref{eq:running_variance}, but for $i \geq k$, $i$ is replaced with the constant
$k$ in the two equations. This provides a way to directly control the adaptation speed of the response characteristics, where smaller $k$ means faster adaptation, and vice versa.

Another possibility is to use the exponentially weighted moving average and variance \parencite{Finch--2009}, which are calculated as

\begin{equation}\label{eq:exp_running_avg}
\mu_{i} = \mu_{i-1} + \alpha(x_{i} - \mu_{i-1}),
\end{equation}

\begin{equation}\label{eq:exp_running_var}
s_{i} = (1 - \alpha)(s_{i-1} + \alpha(x_{i} - \mu_{i-1})^2),
\end{equation}

where $\alpha$ $\in(0,1)$ is an adjustable parameter\footnote{A `forgetting rate' by analogy with the `learning rate' parameter used in backpropagation.}.
In this way, the contribution of older values to the average decreases exponentially with time.

\section{Experimental Verification}
The proposed method allows NNs to adapt to the statistics of the input. Therefore, it is naturally suited for applications such as processing streaming data,
where it is necessary to adapt to the incoming data on the fly.
This section presents an experiment conducted in order to verify that the proposed adaptive input conversion method indeed adapts to the underlying statistics
of the input and is capable of detecting deviations from the average value.

Data from the Numenta Anomaly Benchmark (NAB) database \parencite{Lavin_Ahmad--2015} was used for this purpose.
The NAB database contains several dozen examples of real-world streaming data (as well as a few artificial examples) which contain different types of anomalies.
It covers a diverse range of data sources, such as the number of taxi rides in New York in 30-minute intervals and the number of tweets related to specific tech companies, such as Google and IBM,
in five-minute intervals. The benchmark suite currently includes three algorithms -- Numenta's own HTM algorithm, the Skyline algorithm by Etsy, Inc., and the ADVec algorithm developed by Twitter.
All of these algorithms are designed to detect anomalies in streaming data on the fly.

The dataset selected for this experiment represents the temperature of a certain industrial machine in five-minute intervals, collected over the span of about three months.
There are four anomalies in the data (shaded areas in Figs. \ref{fig:proposed_machine_temperature_system_failure} and \ref{fig:others_machine_temperature_system_failure}),
where the second one represents a planned shutdown, the fourth one represents a catastrophic failure of the machine,
and the third one has been identified by experts as a precursor to the fourth one.
This illustrates the usefulness of online anomaly detection: detecting the third anomaly while the machine was operating would have raised an alarm, potentially prevented the imminent failure and
saving the company operating the machine the cost of repair or replacement.

\begin{figure}
	\centering
	\subcaptionbox{\label{fig:IF_neuron}}
	{
		\begin{tikzpicture}
		\begin{axis}%
		[%
		axis x line=bottom,
		axis y line=left,
		ymin=0,
		xmin=0,
		ymax=1.2,
		xmax=15,
		yscale=0.6,
		ytick={1},
		yticklabels={$\theta$},
		xtick={1.4,2.1,3.1,8,8.92},
		xticklabels={},
		xlabel=Spike arrival time,
		ylabel=Depolarisation level,
		samples=100,
		smooth
		]

		\addplot
		[%
		mark=none,
		thin,
		name path=threshold,
		dotted,
		domain=0:15
		] {1};

		\draw [dotted] (axis cs: 1.4,0.0) -- (axis cs: 1.4,0.07);
		\draw [red] (axis cs: 1.4,0.07) -- (axis cs: 2.1,0.07);
		\draw [dotted] (axis cs: 2.1,0.0) -- (axis cs: 2.1,0.123);
		\draw [red] (axis cs: 2.1,0.123) -- (axis cs: 3.1,0.123);
		\draw [dotted] (axis cs: 3.1,0.0) -- (axis cs: 3.1,0.405);
		\draw [red] (axis cs: 3.1,0.405) -- (axis cs: 8.0,0.405);
		\draw [dotted] (axis cs: 8.0,0.0) -- (axis cs: 8.0,0.84);
		\draw [red] (axis cs: 8.0,0.84) -- (axis cs: 8.92,0.84);
		\draw [dotted] (axis cs: 8.92,0.0) -- (axis cs: 8.92,0.92);
		\draw [red] (axis cs: 8.92,0.92) -- (axis cs: 10.73,0.92);
		\draw [red] (axis cs: 10.73,1) -- (axis cs: 10.73,0);
		\draw [red] (axis cs: 10.73,0) -- (axis cs: 15,0);

		\end{axis}

		\end{tikzpicture}
	}

	\subcaptionbox{\label{fig:LIF_neuron}}
	{
		\begin{tikzpicture}
		\begin{axis}%
		[%
		axis x line=bottom,
		axis y line=left,
		ymin=0,
		xmin=0,
		ymax=1.2,
		xmax=15,
		yscale=0.6,
		ytick={1},
		yticklabels={$\theta$},
		xtick={1.4,3.4,6.2,7.1,7.9,8.4,9.5},
		xticklabels={},
		xlabel=Spike arrival time,
		ylabel=Depolarisation level,
		samples=100,
		smooth,
		clip=false
		]

		\addplot
		[%
		mark=none,
		thin,
		name path=threshold,
		dotted,
		domain=0:15
		] {1};

		\draw [dotted] (axis cs: 1.4,0.0) -- (axis cs: 1.4,0.14);
		\addplot [red,domain=1.4:3.4] {0.14*exp(-(x - 1.4)/3)};
		\draw [dotted] (axis cs: 3.4,0.0) -- (axis cs: 3.4,0.67);
		\addplot [red,domain=3.4:6.2] {0.67*exp(-(x - 3.4)/3)};
		\draw [dotted] (axis cs: 6.2,0.0) -- (axis cs: 6.2,0.83);
		\addplot [red,domain=6.2:7.1] {0.81*exp(-(x - 6.2)/3)};
		\draw [dotted] (axis cs: 7.1,0.0) -- (axis cs: 7.1,0.91);
		\addplot [red,domain=7.1:7.9] {0.91*exp(-(x - 7.1)/3)};
		\draw [dotted] (axis cs: 7.9,0.0) -- (axis cs: 7.9,0.93);
		\addplot [red,domain=7.9:8.4] {0.93*exp(-(x - 7.9)/3)};
		\draw [dotted] (axis cs: 8.4,0.0) -- (axis cs: 8.4,0.95);
		\addplot [red,domain=8.4:9.5] {0.95*exp(-(x - 8.4)/3)};
		\draw [dotted] (axis cs: 9.5,0.0) -- (axis cs: 9.5,0.97);
		\addplot [red,domain=9.5:10] {0.97*exp(-(x - 9.5)/3)};
		\draw (axis cs: 10,1) -- (axis cs: 10,-0.1);
		\addplot [red,domain=10:15] {-0.1*exp(-(x - 10)/3)};

		\end{axis}

		\end{tikzpicture}
	}

	\caption[LIF neuron activation]{A histogram of the depolarisation level of (a) IF and (b) LIF neurons.
		After each spike, an IF neuron maintains its depolarisation indefinitely, allowing it to respond to spikes which arrive far apart in time.
		In contrast, after each incoming spike, the membrane of a LIF neuron slowly repolarises following an exponential law of the form $exp(-t/\tau)$, where
		$\tau$ (the time constant of the neuron) controls the decay rate. Repolarisation continues until the resting potential is reached (here, represented by the $x$ axis).
		Therefore, enough spikes have to arrive within a short amount of time in order to reach the threshold, which is why LIF neurons are useful for detecting synchronised spikes.
		After reaching the threshold, the neuron spikes, and the potential is reset to the resting potential (IF neurons) or slightly
		\textit{below} the resting potential (LIF neurons) in a mechanism known as hyperpolarisaton.
		In general, LIF neurons approximate the dynamics of biological neurons more accurately than IF neurons do.
	}
	\label{fig:IF_and_LIF}
\end{figure}
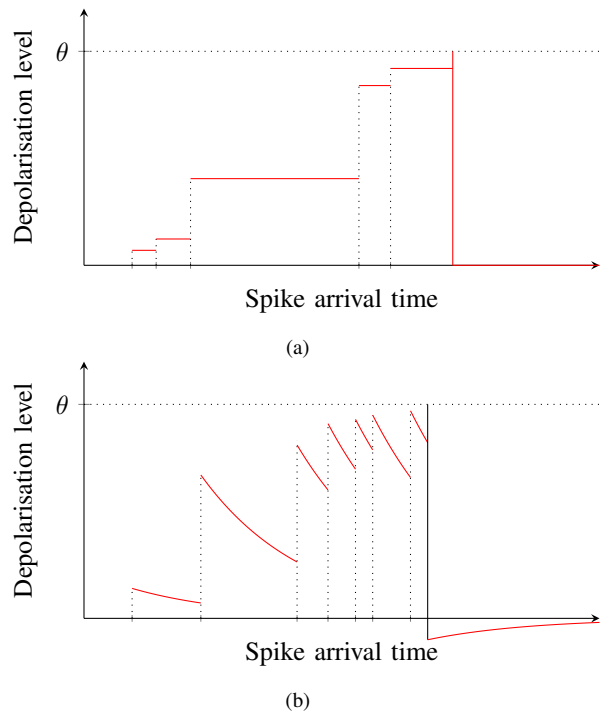

The objective is to detect the anomalies (true positives; TP) as early as possible while minimising the number of false positives (FP) and false negatives (FN),
where FN means a failed anomaly detection.
The spiking NN used in the simulation was composed of 10 input neurons and a single output neuron.
There were no hidden neurons.
Since there was only one variable to monitor, the data points were fed into the network sequentially, where the first neuron received the most recent data point,
the second one received the preceding one and so forth, for up to 10 data points.
The output neuron was a leaky integrate-and-fire (LIF) neuron (Fig. \ref{fig:IF_and_LIF}) with a firing threshold $\theta = 40~mV$ (relative to a resting potential of $0~mV$), a decay constant $\tau = 10~ms$, and a maximal firing rate $\rho = 0.5~spikes/ms$.
All synaptic weights were set to $1$. The temporal distance between successive data points in the data file is $5~min$, which is clearly too long for processing with a spiking NN.
Therefore, this interval was mapped to $10~ms$ of simulation time in order to make it comparable in magnitude to the time constant $\tau$.
A spike produced by the output neuron was recorded as a detected anomaly.

The adaptive conversion scheme was implemented using the exponential running average and variance (Eqs. \ref{eq:exp_running_avg} and \ref{eq:exp_running_var}).
To find the optimal value for $\alpha$, the anomaly detection was performed for $\alpha$ from $0.0005$ to $0.05$ in increments of $0.0005$, and the TP, FP and FN values were recorded for each run.
A simple scoring metric was employed whereby a TP was worth $10$ units, a FP was worth $-1$ unit and a FN was worth $-10$ units, and the total score for a run was calculated by summing all values for TP, FP and FN.
The rationale behind this scoring scheme followed the argument presented in \parencite{Lavin_Ahmad--2015},
namely that investigating a false alarm (FP) is much less expensive than repairing or replacing the machine after an unforeseen failure (FN).
Therefore, failing to detect an anomaly was penalised more heavily than raising a false alarm.

\begin{figure}
	\centering
	\includegraphics[width=\linewidth]{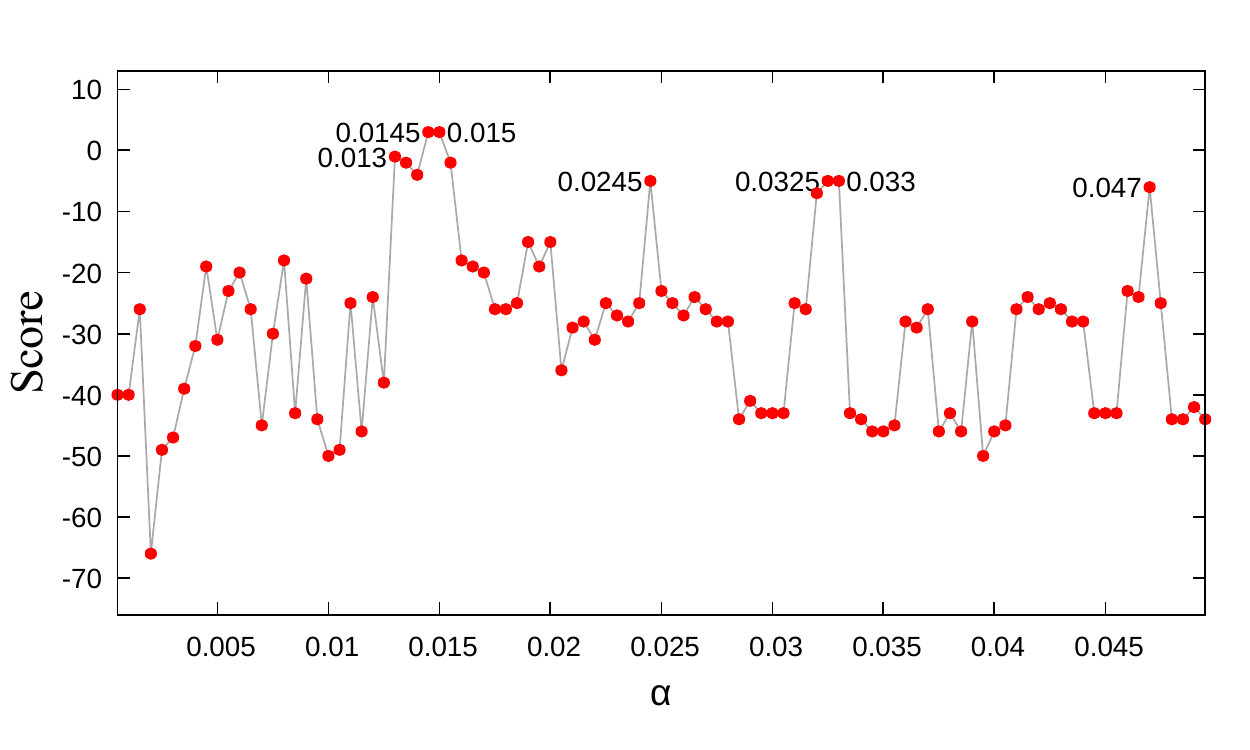}
	\caption[Score per alpha]{Score obtained for different values of the parameter $\alpha$ in Eqs. \ref{eq:exp_running_avg} and \ref{eq:exp_running_var}.
		Although the dependence is highly nonlinear and indicates the presence of many FP and/or FN points, there are seven clearly identifiable peaks, whose corresponding $\alpha$ values are shown in the plot.
		The anomalies detected with $\alpha = 0.013$ and $0.015$ are shown in Fig. \ref{fig:proposed_machine_temperature_system_failure}}.
	\label{fig:score_per_alpha}
\end{figure}

The scores obtained for different values of $\alpha$ are plotted in Fig. \ref{fig:score_per_alpha}. For most values of $\alpha$, the network produced a lot of FN and FP, and accordingly the scores are generally low.
However, there are several peaks, which were investigated further. Upon plotting the anomalies detected for these seven values of $\alpha$, it was found that the network clearly failed to detect the fourth anomaly for
$\alpha=0.0245$, $0.0325$, $0.033$ and $0.047$ and the third anomaly for $\alpha=0.0245$, $0.0325$ and $0.033$. Furthermore, the plots for $\alpha=0.0145$ and $0.015$ were very similar.
Therefore, only the anomalies detected for $\alpha=0.013$ and $0.015$ are shown in Fig. \ref{fig:proposed_machine_temperature_system_failure}.
The results for the same dataset obtained with the Numenta HTM, Etsy Skyline and Twitter ADVec algorithms are shown in Fig. \ref{fig:others_machine_temperature_system_failure}.

\begin{figure}
\centering
	\subcaptionbox{\label{fig:alpha_13}}
	{
		\includegraphics[width=0.9\linewidth]{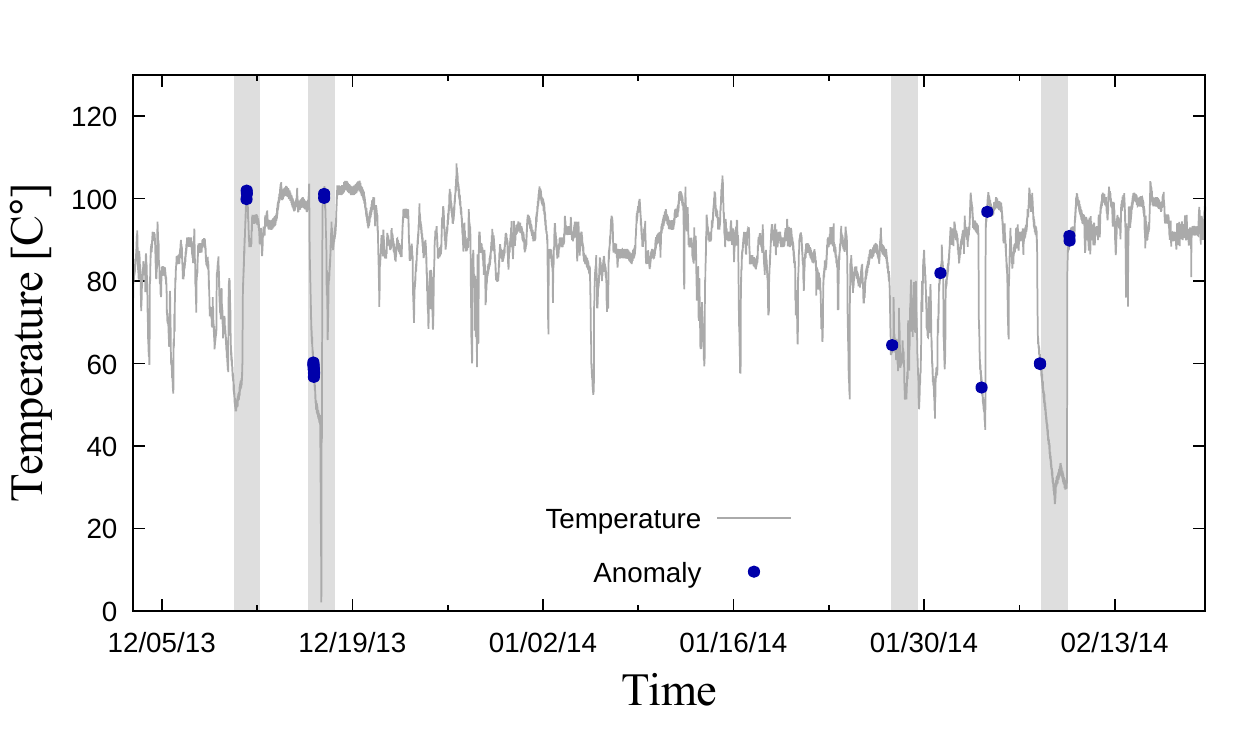}
	}
	\subcaptionbox{\label{fig:alpha_15}}
	{
		\includegraphics[width=0.9\linewidth]{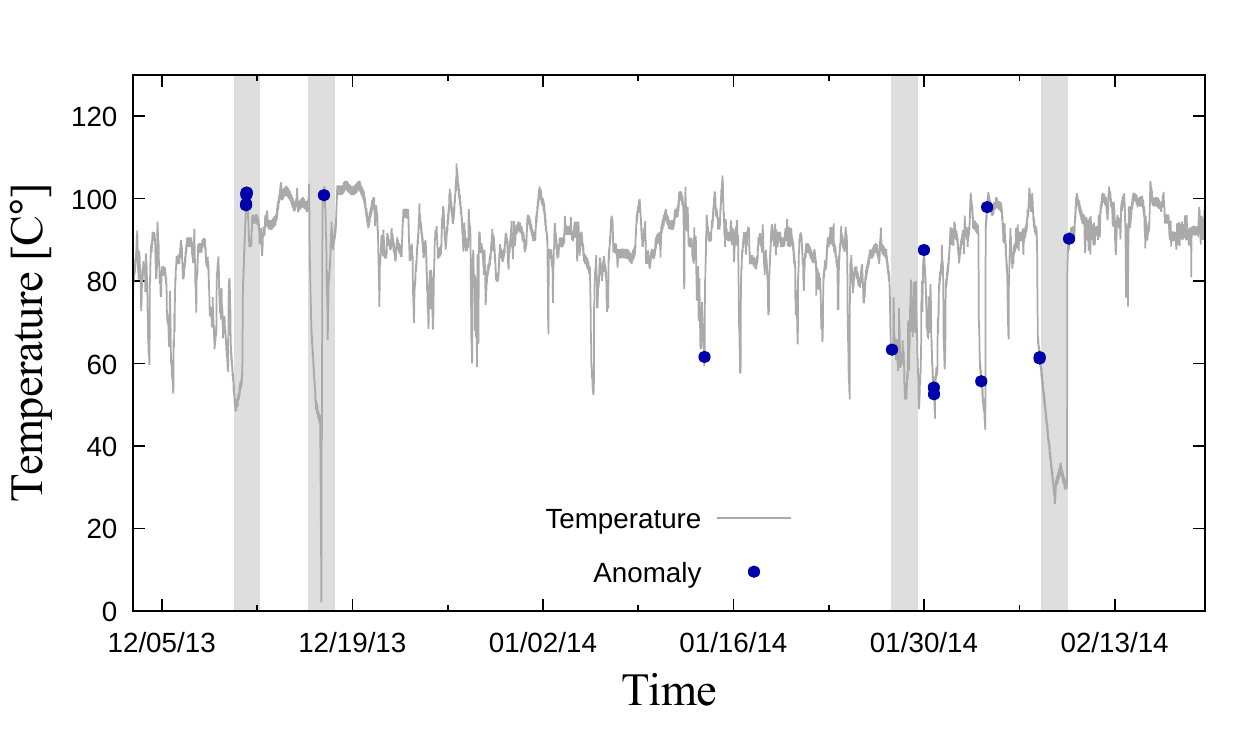}
	}
	\caption[Anomaly detection results with the proposed method]{Anomalies detected with the proposed method with (a) $\alpha = 0.013$ and (b) $\alpha = 0.015$.}
	\label{fig:proposed_machine_temperature_system_failure}
\end{figure}

\begin{figure}
	\centering
	\subcaptionbox{}
	{
		\includegraphics[width=0.9\linewidth]{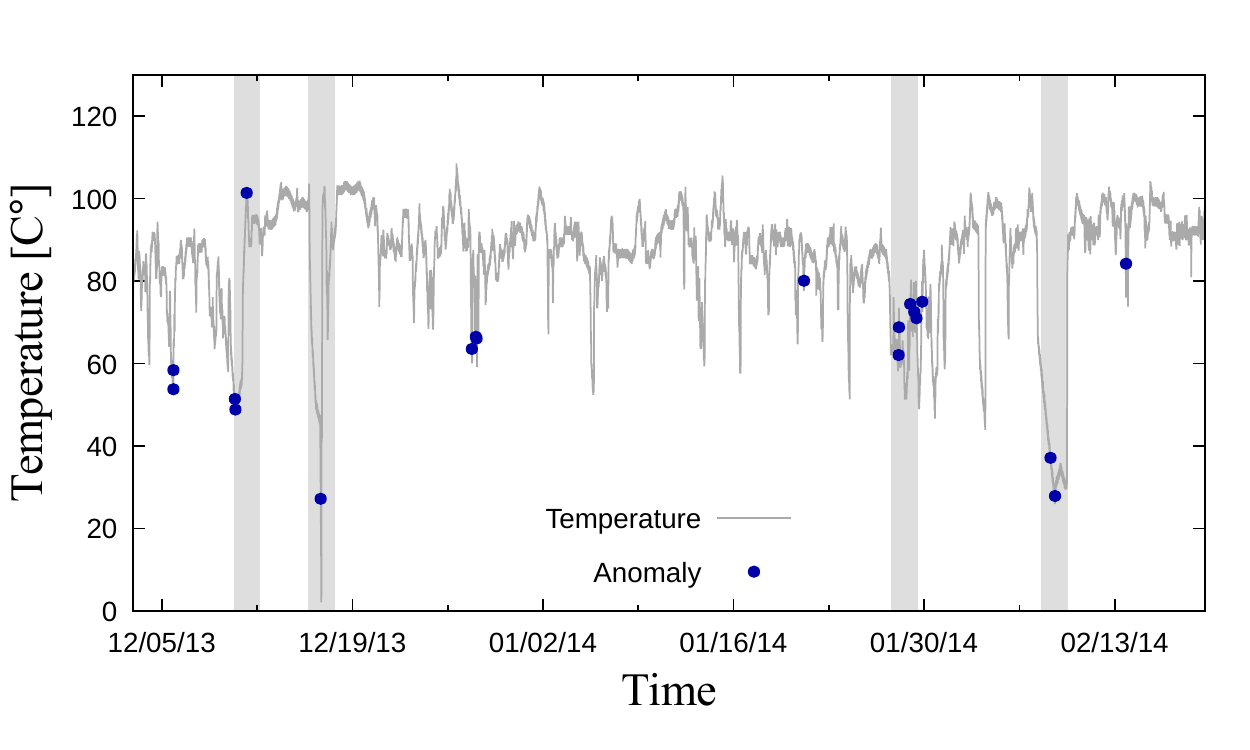}
	}
	\subcaptionbox{}
	{
		\includegraphics[width=0.9\linewidth]{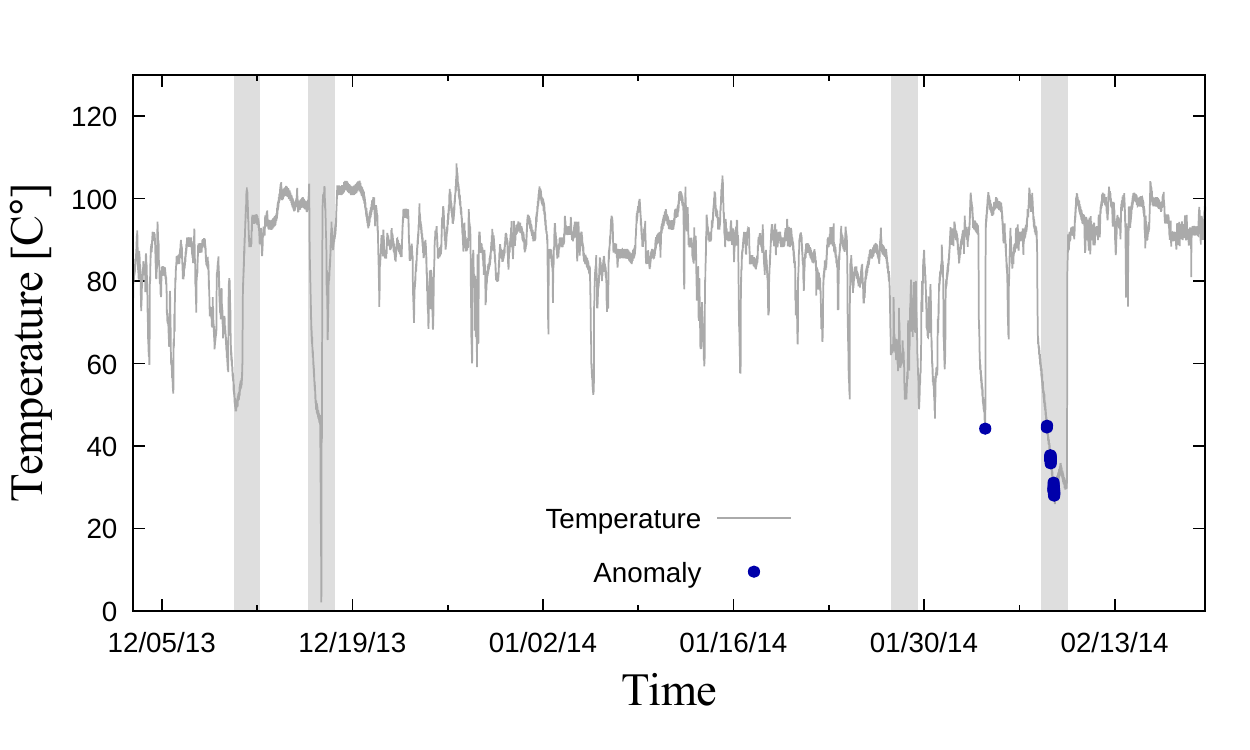}
	}
	\subcaptionbox{}
	{
		\includegraphics[width=0.9\linewidth]{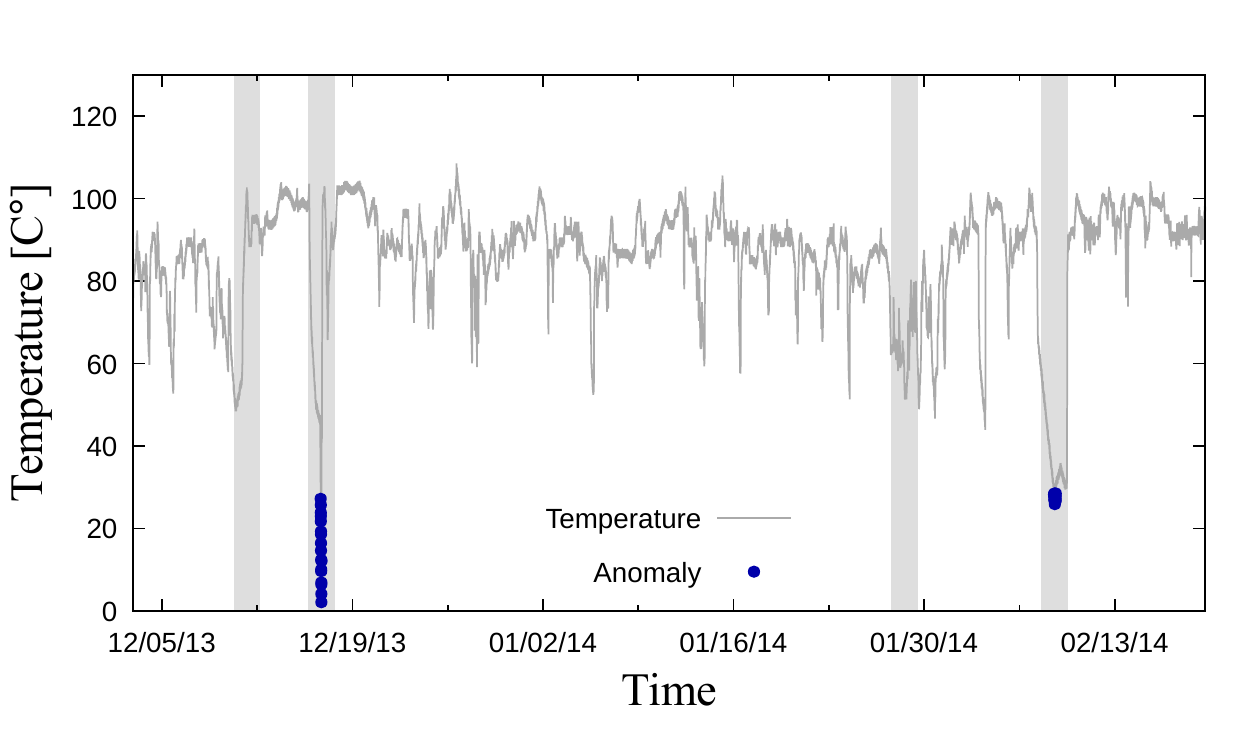}
	}
	\caption[Anomaly detection results with HTM, Skyline and ADVec]{Anomalies detected with (a) Numenta HTM, (b) Etsy Skyline and (c) Twitter ADVec.
		Only HTM detects the third anomaly, which is considered to be a precursor to the catastrophic failure indicated by the fourth anomaly.}
	\label{fig:others_machine_temperature_system_failure}
\end{figure}

Clearly, the spiking NN equipped with the proposed adaptive input conversion method is capable of detecting all four anomalies.
Most importantly, it detects the third anomaly, which is considered to serve as a warning of the catastrophic failure about to occur shortly afterwards.
It is noteworthy that while the Skyline algorithm does not detect an anomaly within the third shaded region,
it does identify the large downward spike after that region as an anomaly.
Therefore, it is conceivable that FPs produced after the third anomaly may be indicative of the abnormal behaviour of the machine in the recent past.

Although the results obtained in the experiment above are comparable with those obtained with existing anomaly detection algorithms,
this is not the ultimate objective of the experiment.
Rather, this is a proof-of-concept experiment demonstrating that the input neurons indeed adapt to their input.
For instance, in Fig. \ref{fig:alpha_13}, it is clear that the network did not produce spikes for any of the peaks and valleys in temperature plot
between the second and third anomaly, even though some of them were rather sharp.
Despite this, the spike at the onset of the third anomaly corresponds to a value which is not as extreme as some of the preceding ones,
which is indicative of an adaptive behaviour in the period between the second and third anomalies.

Note that the NAB results in Fig. \ref{fig:others_machine_temperature_system_failure} were obtained by allowing the algorithms
to learn from the first 15\% of the data before being benchmarked,
which was not done in the current experiment.
In fact, the first anomaly falls within this first interval of 15\%, and therefore it is not included in the formal benchmark of
the Numenta HTM, Twitter ADVec and Etsy Skyline algorithms.
In addition, the proposed method was applied to the same dataset several times in order to tune the value for the parameter $\alpha$.
However, according to the NAB requirements, anomaly detection algorithms such as those presented in Fig. \ref{fig:others_machine_temperature_system_failure}
should tune all of its parameters on the fly, without any look-ahead \parencite{Lavin_Ahmad--2015}.
An extension to the present method in which $\alpha$ is adjusted on the fly is currently being developed.
This will allow for a proper comparison with existing anomaly detection algorithms to be performed on the entire NAB database in a future study.

\section{Discussion}
The adaptive input conversion method presented here allows spiking NNs to convert real-valued input into spike trains for further processing.
However, the conversion method is not limited to spiking NNs.
Indeed, the transfer function in Fig. \ref{fig:abs_tanh} produces a continuous output, and therefore should be equally applicable to classical NNs.
In fact, classical NNs would potentially benefit even more from this method because it eliminates the need for pre-processing
(normalising and shifting to $0$ mean) the input data,
which is currently the recommended practice \parencite{LeCun_Bottou_Orr_Muller--2012}.
As mentioned at the beginning of the paper, spiking NNs have been shown to be strictly more powerful in terms of processing capabilities than classical NNs.
However, classical NNs are essentially idealisations of rate-based processing and output a real number representing the instantaneous firing rate,
whereas spiking NNs are incapable of producing a fractional firing rate because the number of spikes must be an integer.
This could give classical NNs an advantage over spiking NNs for certain applications, but this remains to be confirmed.

In this regard, it is possible to convert the input values into spike latencies rather than spiking rates, which is the approach in the case of GRF-based conversion.
Because time is a continuous variable, spike latencies can be computed with arbitrary precision,
eliminating the problem with the discrete nature of rate-based processing with spiking NNs.
However, converting several input variables into spike latencies with the proposed method would tend to detect input values \textit{close to} the mean
rather than values \textit{far from} the mean.
Although this alternative option was not used in the present study,
it could be of use in certain situations, such as detecting \textit{synchronous} input from two different sources.
Note that the algorithm in its current form is not well suited for applications such as pattern recognition.
The reason for this is that the input layer constantly changes its response characteristics, and therefore the output spike patterns
corresponding to two or more similar input patterns could differ drastically depending on the order in which the patterns are presented
and the temporal distance between the patterns.

Another line of research worth pursuing is the interaction between the proposed adaptive input conversion method and Hebbian learning.
As input neurons adapt their spiking rate to the underlying statistics of the input, Hebbian learning could be employed as a learning paradigm which constantly adjusts the response of the remainder of the network to the changed statistics by modifying synaptic weights while the network is operating.

Finally, as mentioned above, the purpose of the above experiment was to show that the conversion method indeed adapts to its input,
rather than to compete directly with other anomaly detection algorithms.
The reason for this is that at present the optimal value of the parameter $\alpha$ must be found heuristically.
However, the method is expected to become competitive for anomaly detection if this optimal value can be determined automatically and adjusted on the fly.
One possible way to do this is to set the parameter $\alpha$ in
Eqs. \ref{eq:exp_running_avg} and \ref{eq:exp_running_var} based on the mean distance between the last several spikes.
A longer distance should translate into higher $\alpha$, and vice versa.
Another possibility is to employ recurrent connections from the output to the input layer, thus providing a `reference' signal
allowing the network to adapt to its own output.
These ideas, as well as the other two points discussed in this section, are currently being investigated.

\section{Conclusion}
This paper introduced a novel method for converting real-valued input into spike trains suitable for processing with spiking NNs.
The method allows input neurons to track the mean and variance of input variables on the fly, making it applicable to processing streaming data, which was
confirmed through a simple proof-of-concept experiment.
Further research will focus on developing a fully automated scheme for setting all necessary parameters automatically, as well as on investigating the interaction
between adaptive input conversion and Hebbian learning.
The applicability of the proposed method to classical NNs will also be explored.




%
\printbibliography

\end{document}